\documentclass[10pt,twocolumn,letterpaper]{article}

\usepackage{cvpr}              % To produce the CAMERA-READY version
\usepackage{graphicx}
\usepackage{amsmath}
\usepackage{amssymb}
\usepackage{booktabs}

\usepackage{url}            % simple URL typesetting
\usepackage{booktabs}       % professional-quality tables
\usepackage{amsfonts}       % blackboard math symbols
\usepackage{nicefrac}       % compact symbols for 1/2, etc.
\usepackage{microtype}      % microtypography
\usepackage{xcolor}         % colors

\usepackage{nccmath}  
\usepackage{bbm}
\usepackage{amsmath}
\usepackage{amssymb}
\usepackage{mathtools}
\usepackage{amsthm}

\usepackage{amsfonts}

\usepackage[flushleft]{threeparttable} % threeparttable
\usepackage{multirow}

\usepackage{wrapfig}

\usepackage[accsupp]{axessibility} 

\definecolor{Note_color}{rgb}{0.0, 0.0, 0.0}

\newcommand{\revise}[1]{\textcolor{Note_color}{#1}}

\usepackage{layout}
\DeclareMathOperator*{\argmin}{arg\,min}

\usepackage{pifont}
\newcommand\mynuma[1]{\ifcase#1 \or \ding{172}\or \ding{173}\or
  \ding{174}\or \ding{175}\or \ding{176}\or \ding{177}%
  \or \ding{178}\or \ding{179}\or \ding{180}\or \ding{181}\else *\fi\relax}

\newcommand\mynumb[1]{\ifcase#1 \or \ding{182}\or \ding{183}\or
  \ding{184}\or \ding{185}\or \ding{186}\or \ding{187}%
  \or \ding{188}\or \ding{189}\or \ding{190}\or \ding{191}\else *\fi\relax}

\usepackage[pagebackref,breaklinks,colorlinks]{hyperref}

\usepackage[capitalize]{cleveref}
\crefname{section}{Sec.}{Secs.}
\Crefname{section}{Section}{Sections}
\Crefname{table}{Table}{Tables}
\crefname{table}{Tab.}{Tabs.}

\def\cvprPaperID{10016} % *** Enter the CVPR Paper ID here
\def\confName{CVPR}
\def\confYear{2023}

\newcommand{\METHOD}{Auto-CARD}

\begin{document}

\title{Auto-CARD: Efficient and Robust Codec Avatar Driving \\ for Real-time Mobile Telepresence \vspace{-0.5em}}

\author{Yonggan Fu\textsuperscript{1}\thanks{Work done during an internship at Meta.}, \, Yuecheng Li\textsuperscript{2}, \, Chenghui Li\textsuperscript{2}, \, Jason Saragih\textsuperscript{2}, \\ Peizhao Zhang\textsuperscript{2}, \, Xiaoliang Dai\textsuperscript{2}, \, Yingyan (Celine) Lin\textsuperscript{1} \\
\textsuperscript{1}Georgia Institute of Technology \,
\textsuperscript{2}Meta\\
\small{\{yfu314, celine.lin\}@gatech.edu \, \{yuecheng.li, leo.li, jsaragih, stzpz, xiaoliangdai\}@meta.com} \vspace{-0.5em}
}

\maketitle

\begin{abstract}

Real-time and robust photorealistic avatars for telepresence in AR/VR have been highly desired for enabling immersive photorealistic telepresence. However, there still exists one key bottleneck: the considerable computational expense needed to accurately infer facial expressions captured from headset-mounted cameras with a quality level that can match the realism of the avatar's human appearance. To this end, we propose a framework called \METHOD{}, which \textbf{for the first time} enables real-time and robust driving of Codec Avatars when exclusively using merely on-device computing resources. This is achieved by minimizing two sources of redundancy. First, we develop a dedicated \textbf{n}eural \textbf{a}rchitecture \textbf{s}earch technique called AVE-NAS for \textbf{av}atar \textbf{e}ncoding in AR/VR, which explicitly boosts both the searched architectures' robustness in the presence of extreme facial expressions and hardware friendliness on fast evolving AR/VR headsets.
Second, we leverage the temporal redundancy in consecutively captured images during continuous rendering
and develop a mechanism dubbed LATEX to skip the computation of redundant frames. Specifically, we first identify an opportunity from the linearity of the latent space derived by the avatar decoder and then propose to perform adaptive \textbf{lat}ent \textbf{ex}trapolation for redundant frames. For evaluation, we demonstrate the efficacy of our \METHOD{} framework in real-time Codec Avatar driving settings, where we achieve a 5.05$\times$ speed-up on Meta Quest 2 while maintaining a comparable or even better animation quality than state-of-the-art avatar encoder designs.

\end{abstract}

\section{Introduction}
\label{sec:intro}

Enabling immersive real-time experiences has been the key factor in driving the advances of Augmented- and Virtual-Reality (AR/VR) platforms in recent years. 
Photorealistic telepresence~\cite{Stephen18, Shih-En19, schwartz2020eyes, ma2021pixel} is emerging as a technology for enabling remote interactions in AR/VR that aims to impart a compelling sense of co-location among participants in a shared virtual space. One state-of-the-art (SOTA) approach, coined Codec Avatars~\cite{Stephen18}, is comprised of two components: (1) an encoder, which estimates a participant's behavior from sensors mounted on an AR/VR headset, and (2) a decoder, which re-renders the aforementioned behavior to the other parties' headset display using an avatar representation. Both the SOTA encoder and decoder designs have leveraged the expressive power of deep neural networks (DNNs) to enable the precise estimation of human behaviors as well as the high fidelity of rendering, which are critical for immersive photorealistic telepresence.

Despite its big promise, one of the main challenges posed by photorealistic telepresence is the competing requirements between ergonomics and computing resources. On the one hand, power, form factor, and other comfort factors strictly limit the available computing resources on an AR/VR headset device. On the other hand, the DNNs used in SOTA Codec Avatars are computationally expensive and require continuous execution during a telepresence call. It is worth noting that the limited computing resource on an AR/VR device must additionally be shared with other core AR/VR workloads, such as the SLAM-based tracking service, controller tracking, hand tracking, and environment rendering. Therefore, it is highly desirable and imperative to minimize the computation overhead and resource utilization of Codec Avatars, while not hurting their precise estimation of human behaviors and rendering fidelity. This has become a bottleneck limiting their practical and broad adoption.

To close the above gap towards real-time Codec Avatars on AR/VR devices, existing work has focused on reducing the computational cost of the decoder. For example, PiCA~\cite{ma2021pixel} leverages the compute characteristics of modern DSP processors to enable simultaneously rendering up to five avatars on a Meta Quest 2 headset~\cite{ma2021pixel}. On the other hand, efficient encoder designs that can fit the AR/VR computing envelope have been less explored, with most existing works assuming off-device computing scenarios. Specifically, SOTA methods for the encoder such as~\cite{Shih-En19, schwartz2020eyes} are prohibitively heavy with $\sim$3 Giga-floating-point-operations (GFLOPs) for  encoding merely from one image, which is too costly to be continuously executed on SOTA AR/VR headsets. 
Although cloud-based solutions have been explored as an alternative for other AR/VR use cases, on-device encoder processing for Codec Avatars is particularly desired for telepresence applications as a way to better protect the privacy and overcome internet bandwidth limitations.

In this work, we aim to enable real-time encoder inference for Codec Avatars on AR/VR devices. Specifically, the encoder takes in image data captured from headset-mounted cameras (HMC) and outputs facial expression codes for a Variational Auto-Encoder (VAE)~\cite{VAE}, which is used as a decoder following prior works~\cite{Stephen18,Gabriel20}. This target problem is particularly challenging due to two reasons. First, naively reducing the encoder capacity, e.g., by compressing the encoder models to have fewer channels and/or shallower layers, typically results in accuracy degradation, especially for extreme expressions at the tail ends of the data distribution which are often precisely the expressions that contain the most informative social signal. Second, since hardware backends are still nascent for AR/VR use cases, heuristics for hardware-specific optimization may quickly become obsolete. For example, the Qualcomm Snapdragon 865  system-on-a-chip (SoC)~\cite{qualcomm} on Meta Quest 2 headsets and customized accelerators~\cite{sumbul2022system} exhibit different latency/energy constraints. As such, it is important for our developed techniques to be able to automatically adapt to different hardware backends for ensuring their practical and wide adoption, instead of relying on manual optimization strategies that require costly laboring efforts.

To tackle the aforementioned challenges, we develop a framework, dubbed \METHOD{}, for enabling efficient and robust real-time \textbf{C}odec \textbf{A}vata\textbf{r} \textbf{d}riving. \METHOD{} \textit{automatically} minimizes two sources of redundancy in the encoding process of SOTA solutions: architectural and temporal redundancy. We summarize our contributions as follows:
\begin{itemize}
    \vspace{-0.5em}
    \item Our proposed framework, \METHOD{}, is \textbf{the first} method that has enabled real-time  and robust driving of Codec Avatars in AR/VR, exclusively using merely on-device computing resources. 
        \vspace{-0.3em}
    \item \METHOD{} integrates a neural architecture search technique that is tailored for \textbf{av}atar \textbf{e}ncoding (AVE-NAS), minimizing potential model redundancy while explicitly accounting for the fast-evolving hardware design trends of AR/VR headsets. AVE-NAS comprises three NAS components: \underline{(1)} a view-decoupled supernet for enabling distributed near-sensor encoding, \underline{(2)} a hybrid differentiable search scheme for an efficient and effective joint search, and \underline{(3)} an extreme-expression-aware search objective.
        \vspace{-0.3em}
    \item To further reduce temporal redundancy towards real-time encoders for Codec Avatars on AR/VR devices, \METHOD{} additionally integrates a mechanism, dubbed LATEX, to skip the computation of redundant frames. Specifically, we first identify an opportunity from the linearity of the latent space determined by the avatar decoder and then propose to perform adaptive \textbf{lat}ent \textbf{ex}trapolation for redundant frames. 
        \vspace{-0.3em}
    \item Extensive experiments on real-device measurements using AR/VR headsets, i.e., Meta Quest 2~\cite{quest2}, show that our method can achieve a 5.05$\times$ speed-up while maintaining a comparable or even better accuracy than SOTA avatar encoder designs.    
        \vspace{-0.3em}
\end{itemize}

\section{Related Work}
\label{sec:related_work}

\textbf{Codec Avatars.}
Traditional methods for photorealistic human face modeling~\cite{alexander2009digital,seymour2017meet} rely on accurate but complex 3D reconstruction processes, which are not suitable for real-time applications. To enable photorealistic telepresence,~\cite{lombardi2018deep} uses a deep appearance model in a data-driven manner, which has been dubbed a Codec Avatar. It adopts a conditional variational auto-encoder~\cite{kingma2013auto} to jointly model both the face geometry and appearance, where the encoder encodes the facial behavior into latent codes, which are then decoded back to the facial mesh and view-dependent texture by the decoder. To enhance the rendering quality of gaze and eye contact, which are crucial for immersive face-to-face interactions,~\cite{schwartz2020eyes} explicitly models human eyes' geometry and appearance on top of~\cite{lombardi2018deep}. Later works further enhance the photorealism~\cite{Shih-En19,Chu2020} of the rendered avatars or enable more diverse functionalities~\cite{bi2021deep,richard2020audio,xiang2022dressing,bagautdinov2021driving}. For example,~\cite{Shih-En19} leverages the constraints imposed by the multi-view geometry to establish precise correspondence between HMC images and avatars and~\cite{Chu2020} replaces the holistic models with a learned modular representation to enhance the robustness of facial expressions;
\cite{bi2021deep} further extends the model to support novel lighting environments and~\cite{richard2020audio} animates the face model using audio and/or eye tracking.

Despite promising advances in Codec Avatar representation, relatively few studies have been dedicated to on-device generation.  
PiCA~\cite{ma2021pixel} has pioneered efficient decoder design, but it does not address on-device encoding. While PiCA has allowed real-time performance, the strict computational constraints on the device leave little room for encoding and other workloads as decoding five avatars using PiCA will drain out almost all resources on Meta Quest 2~\cite{ma2021pixel}.
Our work is the first to directly address the complexity of on-device encoding to enable real-time Codec Avatar driving.

\textbf{Neural architecture search.}
NAS~\cite{zoph2016neural}, a sub-field of AutoML~\cite{hutter2019automated}, automates the design of optimal DNN architectures in a data-driven manner instead of relying on manually hand-crafted heuristics. It has been applied to many different domains~\cite{tan2019efficientnet, tan2019mnasnet, howard2019searching, liu2018darts,chen2018searching, liu2019auto, chen2019fasterseg,fu2020autogan,lee2020journey,chen2020adabert}.
Early NAS works target SOTA task accuracy at the cost of prohibitive search time~\cite{zoph2016neural, zoph2018learning, real2019regularized} and later innovations adopt weight sharing~\cite{pham2018efficient, bender2018understanding, guo2020single,liu2018darts,cai2019once,yu2020bignas} to greatly reduce the search cost. Amongst these, differentiable NAS~\cite{liu2018darts} is particularly effective as both the model weights and architectures are updated differentiably, greatly boosting search efficiency. In parallel, hardware-aware NAS~\cite{tan2019mnasnet, howard2019searching, tan2019efficientnet,wu2019fbnet, wan2020fbnetv2, cai2018proxylessnas,zhang2021g,fu2021auto} explicitly accounts for model efficiency during the search process.

Although NAS appears to be a good match for optimizing encoder performance for on-device computing, we find that a straightforward application results in degraded  accuracy, especially for extreme and uncommon facial expressions, which are rich in social signals. Additionally, off-the-shelf NAS-optimized architectures from existing works are not optimized for AR/VR hardware. Therefore, we develop a differentiable NAS technique dedicated to the design of Codec Avatar encoders by explicitly incorporating both extreme-expression awareness and hardware awareness. 

\section{Preliminaries about Codec Avatars}
\label{sec:background}

\subsection{Codec Avatars: Inference}
\label{sec:background_inference}

Our work is built on top of~\cite{schwartz2020eyes}, a SOTA Codec Avatar model which explicitly models both the geometry and appearance of eyes to achieve immersive eye contact. 
Specifically, the HMC captured images on AR/VR headsets are first encoded into latent codes $z$, gazes $g$, and key points $y$ \footnote{Although key points are not used by the decoder, they serve as auxiliary supervisions for encoder training and can be used by other AR/VR tasks.}. Next, $z$ and $g$ are transmitted to the receiver end and decoded into face geometry $G_f$, view-dependent texture $T_f$, and their counterparts for the explicit eyeball model, $G_e$ and $T_e$, using separate decoders $D_f$ and $D_e$:
\begin{align}
	\hspace{-0.1em} [G_f, T_f] = D_f(z,v,g) \, , \,\, [G_e, T_e] = D_e(g,v,e(G_f)),
	\label{eq:decoder}
\end{align}
\noindent where $v$ is the view directions and $e$ extracts the vertex positions of the eyelids based on the face geometry $G_f$. The final avatar is rasterized by a differentiable renderer $R$, i.e., $I = R([G_e, G_f], [T_e, T_f])$. For more details, we strongly refer the readers to~\cite{schwartz2020eyes}.

\subsection{Codec Avatars: Training}
\label{sec:background_training}

Simultaneously capturing data for building Codec Avatar encoders and decoders is challenging. This is because a participant's face is occluded from outside-in cameras while wearing a headset. In addition, the wearing of a headset adds additional effects, such as pressure and restrictions on the hairstyle as well as shading effects, that are not desirable for avatar modeling purposes. As such, existing works~\cite{lombardi2018deep,schwartz2020eyes} first train a face (and eyeball) decoder using a headset-free capture in a multi-view outside-in camera system, and then perform a second capture while wearing the headset. Meanwhile, correspondences between the HMC images and the decoder's latent space are found by an inverse-rendering process with a domain-transfer component to account for the differences in lighting and spectra between the HMC images and the outside-in camera images used to build the decoder~\cite{schwartz2020eyes}. A real-time encoder can then be trained to map a subset of the HMC images to their corresponding latent codes, gaze, and key points.

\section{The Proposed \METHOD{} Framework}
\label{sec:method}

\begin{figure*}[t!]
\centering
\vspace{-2.5em}
\includegraphics[width=\linewidth]{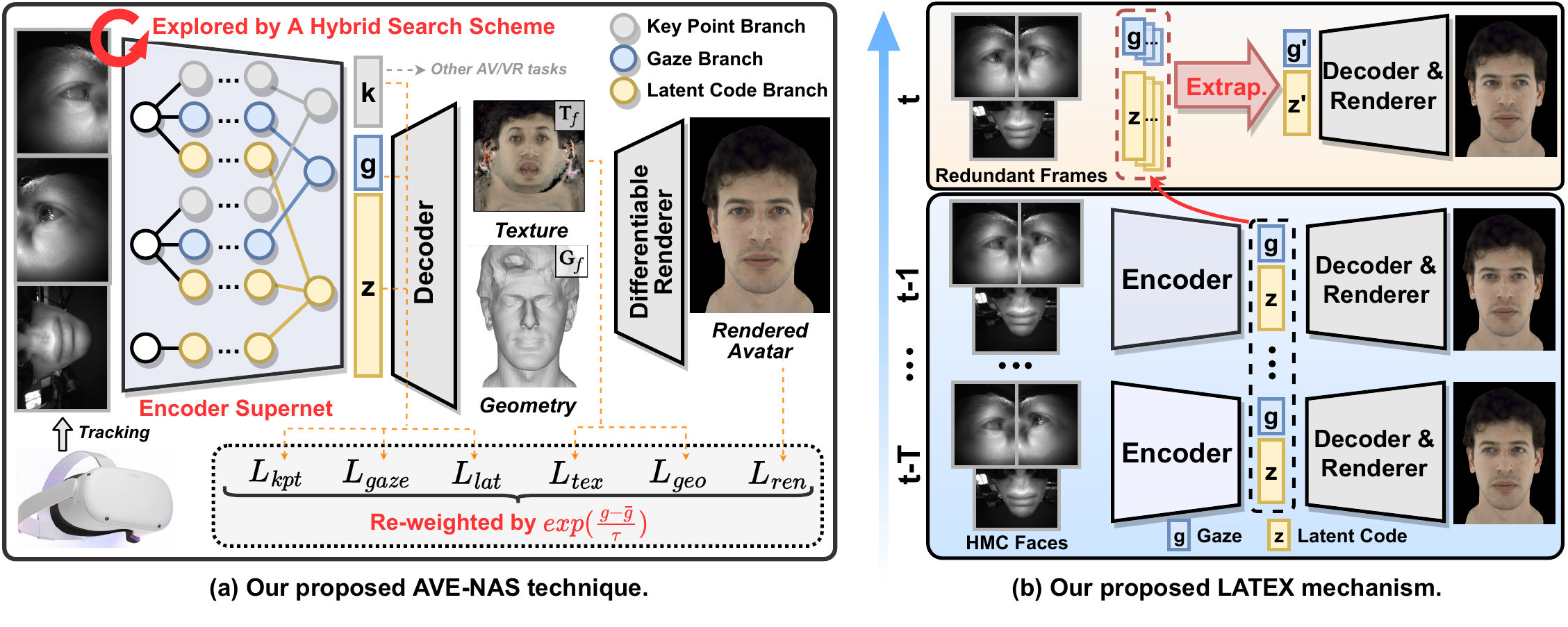}
\vspace{-2.5em}
\caption{An overview of our \METHOD{} framework, integrating the proposed (a) AVE-NAS and (b) LATEX techniques for minimizing the model and temporal redundancy, respectively.}
\label{fig:framework_overview}
\vspace{-1.5em}
\end{figure*}

\subsection{Framework Overview}

The prohibitive computational cost of existing Codec Avatar encoding models is attributed to both the model complexity of per-encoding inference (i.e. model redundancy) and the high encoding frequency for ensuring continuous rendering during remote telepresence (i.e. temporal redundancy). Our method, \METHOD{}, aims to \textit{automatically} minimize these two sources of redundancy while maintaining accuracy via two components: \underline{(1)} AVE-NAS for discovering an efficient and effective encoder architecture in Sec.~\ref{sec:ave_nas} and \underline{(2)} LATEX that adaptively skips computation of redundant frames via latent code extrapolation in Sec.~\ref{sec:latex}. An overview of \METHOD{} is shown in Fig.~\ref{fig:framework_overview}.

\subsection{AVE-NAS: Towards Efficient Avatar Encoders}
\label{sec:ave_nas}
    \vspace{-0.3em}
\subsubsection{Unique Challenges for Avatar Encoder Search}
    \vspace{-0.3em}
To ensure search efficiency for enabling fast adaptation to the evolving AR/VR hardware backends, our AVE-NAS adopts a differentiable search scheme. However, directly extending existing differentiable NAS~\cite{liu2018darts} to the avatar encoding process will lead to non-robust encoders, which can achieve a low average error but fail to capture extreme expressions due to the view-collapse issue analyzed in Sec.~\ref{sec:search_objective}. Additionally, it may also result in hardware-unfriendly encoders that do not align with the hardware design trends and characteristics of AR/VR headsets, causing undesired high energy consumption. Therefore, we explicitly enforce both robustness awareness to extreme expressions and hardware-friendliness to AR/VR headset hardware in the three components of AVE-NAS as elaborated below.

    \vspace{-0.3em}
\subsubsection{AVE-NAS: The Search Space}
\label{sec:search_space}
    \vspace{-0.3em}

\textbf{\indent Considering the trend of distributed near-sensor encoding.}
For building the macro-structure of the encoder, one intuitive choice is to transmit different views captured by different HMCs to a central SoC, which will concatenate those views to be jointly processed by the encoder. However, data streaming from peripheral sensors to the central SoC is costly in terms of energy consumption, especially with high frame rates for achieving smooth facial animation~\cite{pinkham2020algorithm}. This can not only pose severe challenges for the limited battery life on AR/VR headsets but also continuously occupy the bus resources on the headset during telepresence. As such, the desired future trend is to adopt a distributed near-sensor encoding of different captured views~\cite{sumbul2022system}, which has inspired and motivated AVE-NAS's supernet design.

\textbf{Macro-structure of our supernet.} Our AVE-NAS constructs a view-decoupled supernet that independently processes each of the captured partial-face images. As shown in Fig.~\ref{fig:framework_overview} (a), we build a three-branch structure for the left and right eyes to generate the gaze, key points, and latent features, respectively. For the mouth view, we adopt only one branch to generate latent features as it is independent of gazes and key points. Finally, the latent features from different views, each of which is a 128-d vector, are concatenated and then regressed into the final latent code. Note that the data movement cost of three 128-d vectors is much smaller than that of transmitting three images (e.g., 192$\times$192 per view in~\cite{Gabriel20}). Considering different cameras here capture different aspects of the facial appearance and motion, the architectures for the aforementioned three views are optimized separately and exhibit different complexity.

\textbf{Searchable factors.} To ensure sufficient flexibility of the encoder architecture, the search space spans operator types, depth, width, and input resolution. In particular, our AVE-NAS supports Fused-MBConv~\cite{tan2021efficientnetv2}, single convolution, or skip connections as potential operator types, driven by their high execution efficiency on AR/VR headsets.

    \vspace{-0.5em}
\subsubsection{AVE-NAS: The Search Algorithm}
\label{sec:search_alg}
    \vspace{-0.3em}
\textbf{\indent Challenges of the differentiable joint search for avatar encoders.} 
Although the operator and channel numbers in our target search space can be differentiably searched via the commonly used reparameterization trick~\cite{wan2020fbnetv2}, the rendering loss is naturally non-differentiable w.r.t. the input resolution. Nevertheless, searchable input resolutions are highly desired for designing avatar encoders, because the structure/texture information captured from different identities often shows diverse complexity which allows using different resolutions for minimizing the overall model complexity. 
While~\cite{wan2020fbnetv2} achieves differentiable resolution search for classification tasks by inserting paddings inside the input images, such a strategy is not applicable for avatar encoders as it will destroy the structure of the captured human faces. To this end, our design aims to differentiably search for input resolutions, together with the operator types and the number of channels, while maintaining the structure information of human faces.

\textbf{Our rationale.} To achieve the target design above, we first formulate the joint search process as learning the sampling distributions $p(\cdot|\theta)$ parameterized by $\theta \in \{\theta^{op}, \theta^{ch}, \theta^{res}\}$:

\vspace{-0.5em}
\begin{align}
	\argmin_{\theta, w} \mathcal{L}(\theta, w) = \mathbb{E}_{p(\alpha|\theta)}[f(\alpha, w)] \label{eq:search_goal},
\end{align}
\vspace{-1em}

\noindent where $\alpha \in \{\alpha^{op}, \alpha^{ch}, \alpha^{res}\}$ is the sampled design from $p(\alpha|\theta)$, $f$ is the objective function, and $w$ denotes the supernet weights. \textit{The core question} is how to estimate the gradient $\Delta \theta$ for updating $\theta$. Inspired by~\cite{grathwohl2017backpropagation}, we find that both reparameterization tricks~\cite{kingma2013auto} and policy gradients~\cite{sutton1999policy} can produce unbiased gradient estimators for $\Delta \theta$ in Eq.~\ref{eq:search_goal}, where the latter does not require $f$ to be differentiable w.r.t $\alpha$ and thus is well-suited for the desired resolution search. Therefore, our AVE-NAS adopts a hybrid differentiable search scheme to integrate both the above estimators.

\textbf{Proposed Method.} To implement the aforementioned rationale, 
for the operator/channel search, we adopt the reparameterization trick from~\cite{kingma2013auto} to estimate the gradients $\Delta \theta^{oc} = \frac{\partial f(\alpha, w)}{\partial \theta^{oc}} = \frac{\partial f(\alpha, w)}{\partial T(\theta^{oc}, \epsilon)} \frac{\partial T(\theta^{oc}, \epsilon)}{\partial \theta^{oc}}$, where $\theta^{oc} \in \{\theta^{op}, \theta^{ch}\}$, $T(\theta, \epsilon)$ is a continuous function, and $\epsilon$ is a random variable. In particular, we formulate the output of each layer as a weighted sum of all candidate choices, e.g., for the operator search, the ($l$+1)-th layer $x_{l+1}=\sum_{i} T(\theta^{op}_{l,i}, \epsilon) \cdot \alpha^{op}_i(x_l)$ where $T(\theta^{op}_{l,i}, \epsilon)=\frac{exp[(\theta^{op}_{l,i} + \epsilon)/\tau]}{\sum_{i} exp[(\theta^{op}_{l,i} + \epsilon)/\tau]}$ is the Gumbel Softmax function~\cite{jang2016categorical}, $\epsilon$ is sampled from the Gumbel distribution, and $\alpha^{op}_i$ is the $i$-th operator. For the channel search, we follow the channel masking strategy in~\cite{wan2020fbnetv2} for search efficiency.

For the resolution search, we adopt policy gradients~\cite{sutton1999policy} to estimate the gradients $\Delta \theta^{res}=f(\hat{\alpha}, w) \cdot \frac{\partial}{\partial \theta^{res}} log \, p(\hat{\alpha}|\theta^{res})$ derived from Eq.~\ref{eq:search_goal}, where $\hat{\alpha} \sim p(\alpha|\theta^{res})$ is the sampled resolution. Here $-f(\hat{\alpha}, w)$ can be viewed as the reward in reinforcement learning~\cite{sutton1999policy, lillicrap2015continuous}. In our design, to make $p$ differentiable w.r.t. $\theta^{res}$, we adopt the Gumbel Softmax function as $p(\alpha|\theta^{res})$ for sampling $\hat{\alpha}$. In addition, to stabilize the search process and reduce the variance in $\Delta \theta^{res}$, we sample $\hat{\alpha}$ once every K iterations and average the corresponding rewards for updating $\theta^{res}$. The searched encoder will be trained from scratch.

\begin{figure}[!t]
\centering
\includegraphics[width=0.95\linewidth]{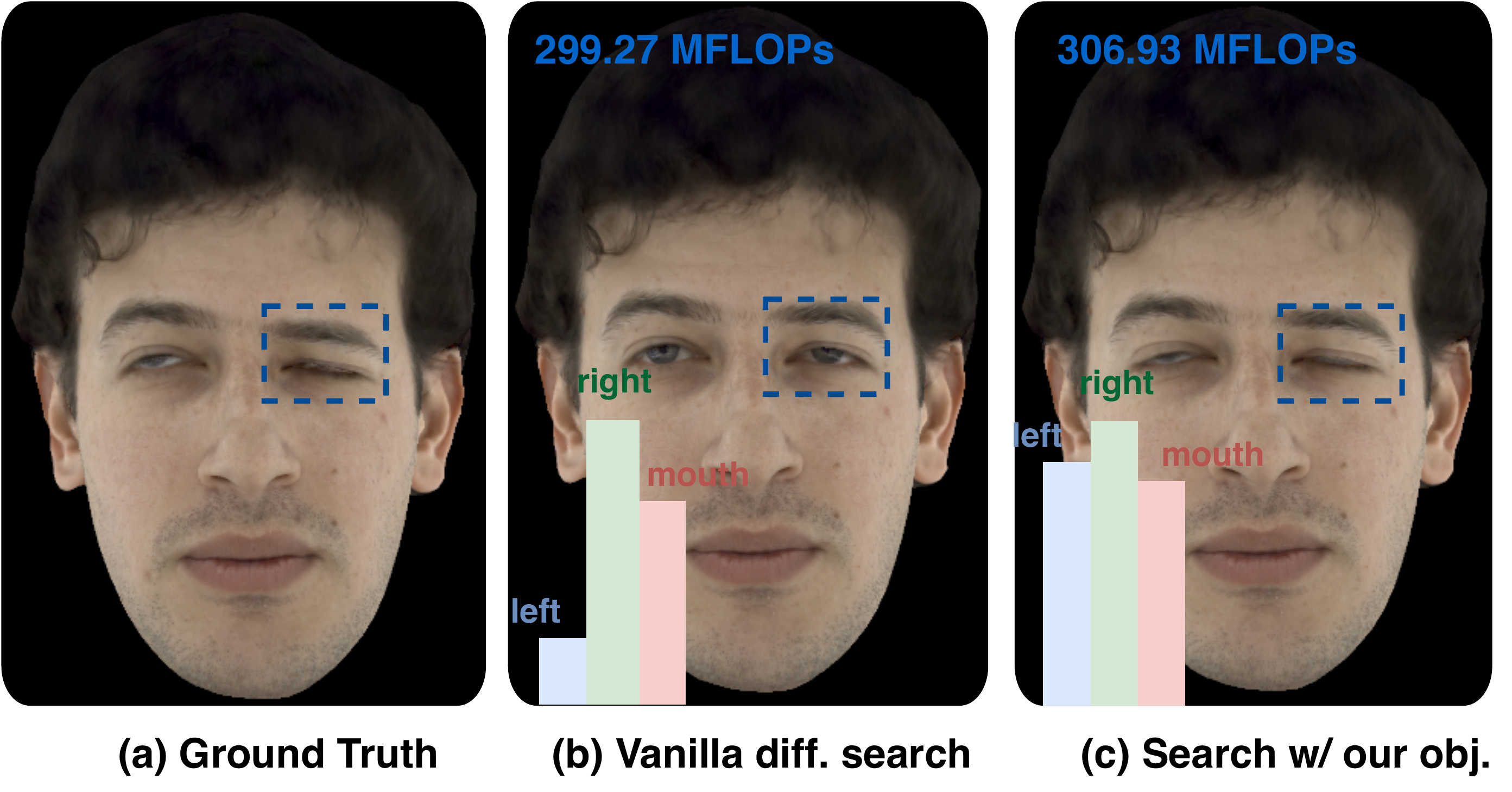}
\vspace{-1em}
\caption{Visualize the rendered avatars by the searched encoders w/o and w/ our proposed objectives, where the normalized FLOPs distributions across three views are annotated.}
\label{fig:view-collapse}
\vspace{-1.5em}
\end{figure}

    \vspace{-1em}
\subsubsection{AVE-NAS: The Search Objective}
\label{sec:search_objective}

\textbf{\indent The view-collapse issue.} 
We empirically find that although NAS-searched encoders can achieve low reconstruction losses on average, they often suffer from notable distortions under extreme expressions, especially those with spatial asymmetry. An example with one eye open and the other closed is shown in Fig.~\ref{fig:view-collapse} (b), where we can see that the avatar fails to precisely reproduce expression. We discover that while leveraging cross-view correlations (e.g., co-occurring movements between the left and right eyes) to optimize the encoder architecture can lead to reduced model capacity and thus an improved latency, doing so under a latency constraint results in too lightweight structures that fail to capture specific views.
The reason is that although those correlations hold for common expressions and thus enable satisfactory average rendering losses, the resulting encoder is non-robust against uncommon and extreme expressions.

\textbf{Proposed extreme-expression-aware objective.} To tackle the issue above, we propose an extreme-expression-aware rendering loss for both the encoder search and training to adaptively re-weight the captured faces based on the rareness of the corresponding expressions. Inspired by the focal loss~\cite{lin2017focal}, we propose to pay higher attention to the rare facial expressions for boosting the robustness of the searched encoder architecture, whereas the vanilla focal loss in~\cite{lin2017focal} is not applicable for rendering tasks like Codec Avatars. To implement our target design, a proper proxy that can indicate the rareness of the expressions is critical.

As pointed by~\cite{schwartz2020eyes}, the rendering quality of gaze and eye contact is the key to achieving immersive face-to-face interactions, while the failure cases with visually unnatural expressions often occur together with inaccurate predictions of the corresponding eye textures. 
This motivates us to adopt the decoded eye textures as a proxy to indicate the expression rareness. Since the decoded eye textures mainly depend on the predicted gaze as introduced in Sec.~\ref{sec:background}, we propose to directly adopt the difference between the predicted gaze of the current face $g$ and the moving averaged gaze $\bar{g}$ of all previous training faces, the latter of which depicts the most common gaze among all the captured faces, to indicate the rareness of the current facial expression. Formally, we formulate our extreme-expression-aware objective $\tilde{\mathcal{L}}$ as:

\begin{align}
	\tilde{\mathcal{L}} = exp(\frac{g-\bar{g}}{\tau}) \cdot \mathcal{L} \, , \,\,\,\,  \bar{g} = m \cdot \bar{g} + (1-m) \cdot g,
	\label{eq:extreme-expression-aware-objective}
\end{align}
\vspace{-1em}

\noindent where we adopt an exponential moving average for performing sample-wise re-weighting. Here, $\tau$ is a temperature parameter that controls the sharpness and $m$ is a momentum factor. Note that this same objective is applied in both the encoder search and training stages to avoid view collapse during search and enhance robustness during training.

\textbf{Detailed loss design.} In addition to using regressing losses for the latent code, gaze, geometry, texture, and key points in~\cite{schwartz2020eyes}, we additionally introduce a rendering loss in the RGB space to search/train the encoders in order to help better refine the high-frequency details of the rendered avatars. This plays a similar role as the perceptual losses in~\cite{johnson2016perceptual}. In particular, the decoded geometry and texture from both the predicted encoding and ground-truth encoding are fed into a differentiable renderer $R$ to rasterize the avatars into RGB images~\cite{schwartz2020eyes} and for calculating their Mean-Square-Error (MSE) loss. Our final loss function can be formulated as follows, where we omit the coefficients for simplicity: 
\vspace{-1em}
\begin{align}
	\mathcal{L} = \mathcal{L}_{latent} + \mathcal{L}_{gaze} + \mathcal{L}_{geo} + \mathcal{L}_{tex} + \mathcal{L}_{kpt} + \mathcal{L}_{ren},
	\label{eq:loss}
\end{align}
\noindent where $\mathcal{L}_{latent}=||\hat{z}-z||_{2}^{2}$ (similarly for $\mathcal{L}_{gaze}$ and $\mathcal{L}_{kpt}$), $\mathcal{L}_{geo}= ||\hat{G_f}-G_f||_{2}^{2} + ||\hat{G_e}-G_e||_{2}^{2}$ (similarly for ${L}_{tex}$), and $\mathcal{L}_{ren}=||R([\hat{G_e}, \hat{G_f}], [\hat{T_e}, \hat{T_f}]) - R([G_e, G_f], [T_e, T_f])||_{2}^{2}$.

\textbf{Enforced latency constraint.} During search, we also enforce a latency constraint $\mathcal{L}_{lat}$, following~\cite{wan2020fbnetv2}, based on real-device measurements on AR/VR headsets to control the efficiency of the searched encoders. In particular, we build a measured latency look-up table  on Meta Quest 2 for the candidate operators in our search space and use the summed-up latency of the sampled operators as that of the whole network, based on the observation that the runtime of each operator is independent of other operators, following~\cite{wu2019fbnet}.

\subsection{LATEX: Leveraging Temporal Redundancy}
\label{sec:latex}
\textbf{Motivation.}
During continuous encoding, modeling temporal correlations between latent codes from consecutive frames can indicate temporal redundancy and thus enable skipping computation for redundant frames. 
As VAEs~\cite{VAE} are designed for smooth latent spaces, if the latent space of Codec Avatar features good linearity, i.e., linear motions in 3D worlds correspond to linear traversal in the latent space, a simple linear extrapolation can derive the latent code of the current frame from that of previous frames.

    \vspace{-1em}
\subsubsection{Evaluating The Linearity of The Latent Space}
    \vspace{-0.3em}

\begin{figure}[!t]
\centering
\includegraphics[width=0.95\linewidth]{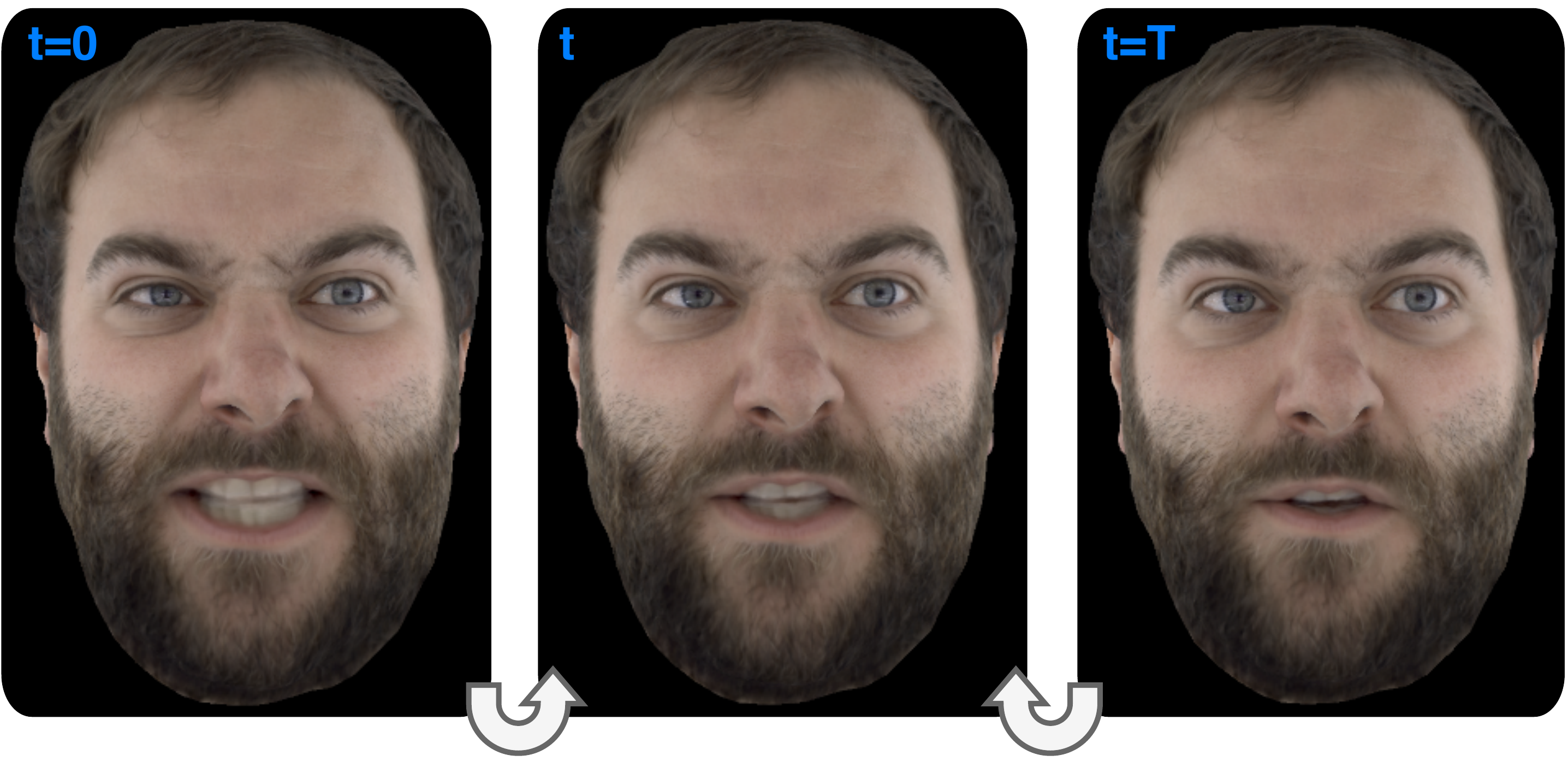}
\vspace{-1em}
\caption{Visualize the rendered avatar (in the middle) decoded from the latent code $z_t$ based on the interpolation of $z_0$ and $z_T$.}
\label{fig:interpolation}
\vspace{-1em}
\end{figure}

We evaluate the linearity of the latent space by linearly interpolating the latent codes of the first and last frames predicted by the encoder, denoted by $z_0$ and $z_T$, respectively, in a batch of size $T$ to approximate the latent codes of all intermediate frames, i.e., $z_t = \frac{T-t}{T} \cdot z_0 + \frac{t}{T} \cdot z_T$. 
For temporal windows of size T=8, we find that this simple approach can achieve a comparable rendering quality and even reduce some jittering effects with an example provided in Fig.~\ref{fig:interpolation}. This indicates that the latent space determined by the decoder indeed features a decent linearity.

    \vspace{-1em}
\subsubsection{Proposed Adaptive Latent Extrapolation}
    \vspace{-0.3em}
Considering that the keyframes, which feature a sudden change in facial expressions and/or motions, cannot be linearly extrapolated based on previous frames, an automatic mechanism that can distinguish the key and redundant frames is highly desired. As such, we propose the LATEX technique to adaptively decide whether to derive the latent code via \underline{(1)} linear extrapolation or \underline{(2)} encoder inference, using a lightweight early prediction mechanism. In particular, we add an extra prediction head (with $<$ 2\% FLOPs overhead) to directly estimate the latent code, which can serve as a proxy for identifying key frames. 
If the difference between the early predicted latent code of the current frame and that of the previous frame is larger than a specified threshold, a complete encoder inference is activated; otherwise, a linear exploration is performed to acquire the latent code (i.e., $z_t = z_{t-1} + \frac{z_{t-1} - z_{t-T}}{T-1}$) based on that of previous $T$ frames, where the predicted gaze and key points are extrapolated in a similar way. In this way, instantaneous trade-offs between the overall latency and rendering quality can be achieved by varying the threshold.

\begin{table}[!t]
\centering
\vspace{-0.5em}
\caption{Benchmark the searched encoders with SOTA encoder designs in terms of measured latency on Quest 2/Pixel 3 and rendering MSE across different identities and views.}
\vspace{-0.5em}
\resizebox{0.98\linewidth}{!}
{    
\begin{tabular}{c|cccc|ccc}
\toprule
\multirow{4}[12]{*}{\textbf{Iden.}} & \textbf{Model} & \textbf{EEM} & \textbf{\begin{tabular}[c]{@{}c@{}}EEM\\ -ch50\end{tabular}} & \textbf{\begin{tabular}[c]{@{}c@{}}EEM\\ -res50\end{tabular}} & \textbf{\begin{tabular}[c]{@{}c@{}}AVE-L \\ (Ours)\end{tabular}} & \textbf{\begin{tabular}[c]{@{}c@{}}AVE-M \\ (Ours)\end{tabular}} & \textbf{\begin{tabular}[c]{@{}c@{}}AVE-S \\ (Ours)\end{tabular}} \\ \cmidrule{2-8}
 & \textbf{MFLOPs} & 2930.77 & 765.38 & 747.44 & 605.14 & 306.93 & \textbf{174.75} \\ \cmidrule{2-8}
 & \textbf{\begin{tabular}[c]{@{}c@{}}Lat. (ms)\\ Quest 2\end{tabular}} & 12.48 & 10.02 & 9.40 & 4.59 & 3.26 & \textbf{2.47} \\ \cmidrule{2-8}
 & \textbf{\begin{tabular}[c]{@{}c@{}}Lat. (ms)\\ Pixel 3\end{tabular}} & 483.47 & 164.27 & 117.27 & 70.53 & 52.61 & \textbf{37.78} \\ \midrule
\multirow{3}{*}{S1} & Front & 8.48 & 8.54 & 11.27 & \textbf{6.91} & 7.46 & 7.54 \\
 & Left & 8.11 & 8.36 & 10.84 & \textbf{6.80} & 7.29 & 7.41 \\
 & Right & 8.04 & 8.07 & 10.65 & \textbf{6.45} & 7.01 & 7.03 \\ \midrule \midrule
\multirow{3}{*}{S2} & Front & 15.70 & 16.08 & 22.77 & \textbf{14.63 } & 15.10 & 16.21 \\
 & Left & 14.52 & 15.08 & 21.08 & \textbf{13.74} & 14.56 & 15.22 \\
 & Right & 17.51 & 18.00 & 24.82 & \textbf{16.17} & 16.31 & 17.53 \\ \midrule \midrule
\multirow{3}{*}{S3} & Front & 12.03 & 12.85 & 15.53 & \textbf{10.91} & 11.42 & 12.25 \\
 & Left & 12.00 & 12.88 & 14.96 & \textbf{10.93} & 11.48 & 11.99 \\
 & Right & 12.73 & 13.62 & 16.45 & \textbf{11.41} & 11.83 & 13.05 \\ \midrule \midrule
\multirow{3}{*}{S4} & Front & 17.42 & 18.71 & 21.40 & \textbf{15.62} & 16.33 & 16.98 \\
 & Left & 19.12 & 20.41 & 23.22 & \textbf{16.80} & 17.57 & 18.01 \\
 & Right & 17.47 & 18.72 & 21.27 & \textbf{15.56} & 16.08 & 16.94 \\ \midrule \midrule
\multirow{3}{*}{S5} & Front & 7.01 & 7.81 & 15.95 & \textbf{5.78} & 5.94 & 6.06 \\
 & Left & 7.32 & 8.05 & 16.09 & \textbf{6.22} & 6.33 & 6.45 \\
 & Right & 7.10 & 7.93 & 15.45 & \textbf{5.92} & 6.03 & 6.24 \\ \midrule \midrule
\multirow{3}{*}{S6} & Front & 19.52 & 20.77 & 25.05 & \textbf{17.34} & 18.39 & 19.08 \\
 & Left & 26.28 & 26.47 & 33.76 & \textbf{22.27} & 24.33 & 24.34 \\
 & Right & 15.89 & 16.73 & 22.25 & \textbf{14.27} & 15.41 & 15.55 \\ \bottomrule
\end{tabular}
}
\label{tab:sota_render_loss}
\vspace{-1.5em}
\end{table}

\begin{figure*}[ht]
\centering
\vspace{-3.5em}
\includegraphics[width=0.8\linewidth]{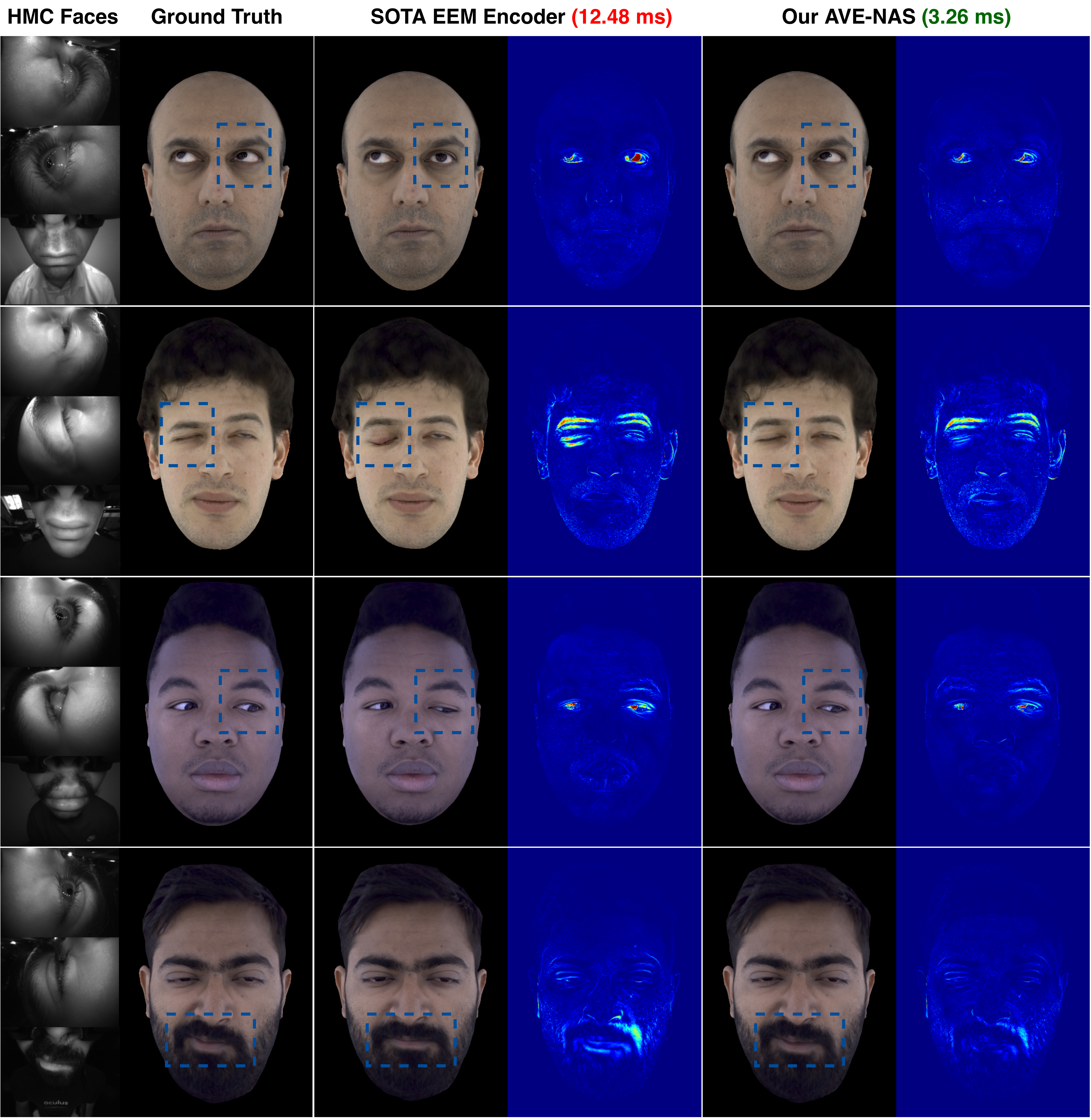}
 \vspace{-0.5em}
\caption{Benchmark the rendering quality achieved by our searched AVE-M against SOTA encoder EEM~\cite{Gabriel20} (zoom-in for better view).}
\label{fig:qualitative}
\vspace{-1.5em}
\end{figure*}

    \vspace{-0.5em}
\section{Experimental Results}
\label{sec:exp}
\vspace{-0.3em}

\subsection{Experiment Setup}
\label{sec:setup}
    \vspace{-0.3em}

\textbf{Dataset.}
We adopt the multiview video dataset captured by the face capture system described in~\cite{lombardi2018deep, schwartz2020eyes}. 
\revise{In particular, we adopt $\sim$12k frames (33 camera views per frame) captured by a large multi-camera capture apparatus for training the decoder and $\sim$42k HMC-captured frames (3 camera views per frame) for searching/training/testing the encoder.
The HMC captured data of each identity consists of different segments featuring specific expressions, facial motions, and speech, among which we randomly select part of the expression segments as testing data and leave other segments as training data.} 
We evaluate our \METHOD{} framework on 10 identities (4 of which are provided in the appendix) that vary in gender, ethnicity, and age. Encoding estimates are evaluated by decoding into 3 different viewing directions: frontal, left, and right views.
The image size of the rendered avatar is set to be 1024$\times$736.

\textbf{Devices and measurement settings.} 
We measure the real-device latency of our searched encoders on two devices with different resource settings, including the SnapDragon 865 SoC~\cite{qualcomm} on the AR/VR headset Meta Quest 2~\cite{quest2} and the Google Pixel 3 mobile phone~\cite{pixel3}. For SnapDragon 865 SoC on Meta Quest 2, we adopt an in-house compiler to optimize the dataflow for mapping the DNN workloads to the hardware; For Pixel 3, we convert PyTorch models to ONNX, which are then compiled to the TFLite format for execution, following~\cite{li2021hw}.

\textbf{Baselines.} We adopt the SOTA encoder design in~\cite{schwartz2020eyes} as our baseline, which is denoted as EEM. In addition, we also adopt two manually compressed encoder designs, which scale the channel numbers or input resolutions by 0.5 on top of EEM and thus denote EEM-ch50/EEM-res50, as our baselines. We benchmark both quantitative reconstruction quality, in terms of the MSE between the rasterized pixels of the rendered avatars and the ground truth, and qualitative rendering quality, in terms of the visual effects under hard expressions. Detailed encoder structures and hyper-parameters of our \METHOD{} are provided in the appendix.

    \vspace{-0.2em}
\subsection{Evaluating AVE-NAS}
    \vspace{-0.2em}

\textbf{Benchmark with SOTA designs.} By default we perform AVE-NAS on one identity with a rich facial motion under different latency constraints, which are denoted AVE-S, AVE-M, and AVE-L, and then generalize the searched encoder structures to other identities if not specifically stated, aiming at validating the generality of the searched encoders. The details of searched structures are provided in the appendix.

\underline{Quantitative comparison:} As shown in Tab.~\ref{tab:sota_render_loss}, we report the MSE between rendered avatars and the ground truth across 6 identities and 3 view directions. We can observe that: \underline{(1)} as compared to the SOTA EEM encoder~\cite{schwartz2020eyes}, our searched encoders consistently achieve better MSE-efficiency trade-offs across identities based on the measurement on Meta Quest 2, e.g., our searched AVE-L achieves a 2.72$\times$ speed-up over EEM while also reducing the rendering MSE by 1.45 on average and our AVE-S achieves a 5.05$\times$ speed-up with an average MSE reduction of 0.46; \underline{(2)} our searched encoders achieve notably better compression effectiveness over manually compressed EEM variants, indicating that jointly exploring multiple dimensions of the encoder architecture is crucial for maintaining high rendering quality.

\underline{Qualitative comparison:}  We visualize the rendered avatars across different identities as well as the corresponding difference maps against the ground truth. As shown in Fig.~\ref{fig:qualitative}, we can observe that our searched encoders can consistently show visually better rendering quality under extreme expressions, i.e., although the SOTA EEM encoder suffers from unnatural facial motions on the eye/mouth regions, our AVE-M can still render photorealistic expressions while achieving a 3.83$\times$ speed-up on Quest 2 headsets.

\begin{table}[h]
\vspace{-0.5em}
\centering
\caption{Benchmark the searched encoders by AVE-NAS and its variant VC-NAS in terms of the rendering MSE.}
\vspace{-0.5em}
\resizebox{0.98\linewidth}{!}
{    
\begin{tabular}{ccccccccc}
\toprule
\textbf{Model} & \textbf{MFLOPs} & \textbf{Lat. (ms)} & \textbf{S1} & \textbf{S2} & \textbf{S3} & \textbf{S4} & \textbf{S5} & \textbf{S6} \\ \midrule
AVE-M & 306.93 & 3.26 & \textbf{7.46} & \textbf{14.63} & \textbf{10.91} & \textbf{16.33} & \textbf{5.94} & \textbf{18.39} \\ \midrule
VC-M & 299.27 & 3.07 & 8.87 & 17.51 & 12.26 & 17.83 & 7.26 & 19.47 \\ \midrule \midrule
AVE-S & 174.75 & 2.47 & \textbf{7.54} & \textbf{16.21} & \textbf{12.25} & \textbf{16.98} & \textbf{6.06} & \textbf{19.08} \\ \midrule
VC-S & 200.05 & 2.53 & 9.37 & 17.97 & 13.17 & 17.97 & 7.17 & 19.66 \\ \bottomrule
\end{tabular}
}
\label{tab:view-collapse}
\vspace{-0.5em}
\end{table}

\begin{figure}[!htb]
\centering
\vspace{-1em}
\includegraphics[width=0.95\linewidth]{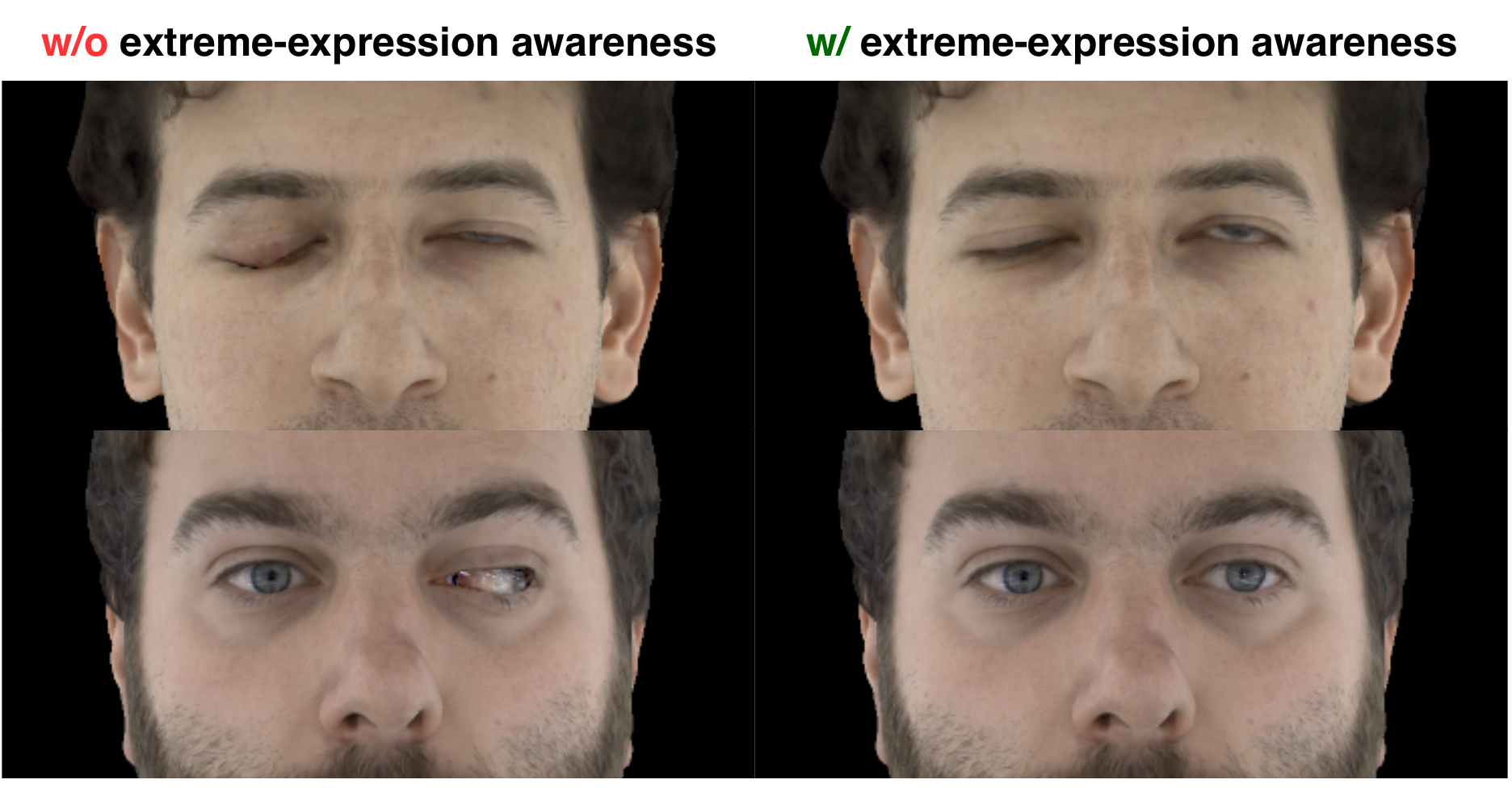}
\vspace{-0.7em}
\caption{Visualize the rendered expressions of the encoders searched w/o and w/ our proposed objective.}
\label{fig:exp_extreme_expression}
\vspace{-1em}
\end{figure}

\textbf{Necessity of extreme-expression-aware objectives.} 
We benchmark the searched encoders by AVE-NAS and its variant without the extreme-expression-aware objective, denoted as VC-NAS since it may suffer from the view-collapse issue analyzed in Sec.~\ref{sec:search_objective}, under the same latency constraint. As shown in Tab.~\ref{tab:view-collapse}, AVE-NAS outperforms VC-NAS in terms of rendering MSE across all identities under a comparable latency on Quest 2. Furthermore, we visualize the rendered avatars under expressions with rare gazes in Fig.~\ref{fig:exp_extreme_expression} and observe that the decoded eye textures of VC-NAS may suffer from great distortions under certain expressions and the rendered eyelids/eyeballs notably lose the fidelity caused by the collapsed left- or right-eye branches. In contrast, after introducing extreme-expression awareness, the accuracy is enhanced significantly.

    \vspace{-0.3em}
\subsection{Evaluating LATEX}
    \vspace{-0.3em}

\begin{figure}[!ht]
\centering
% \vspace{-1em}
\includegraphics[width=\linewidth]{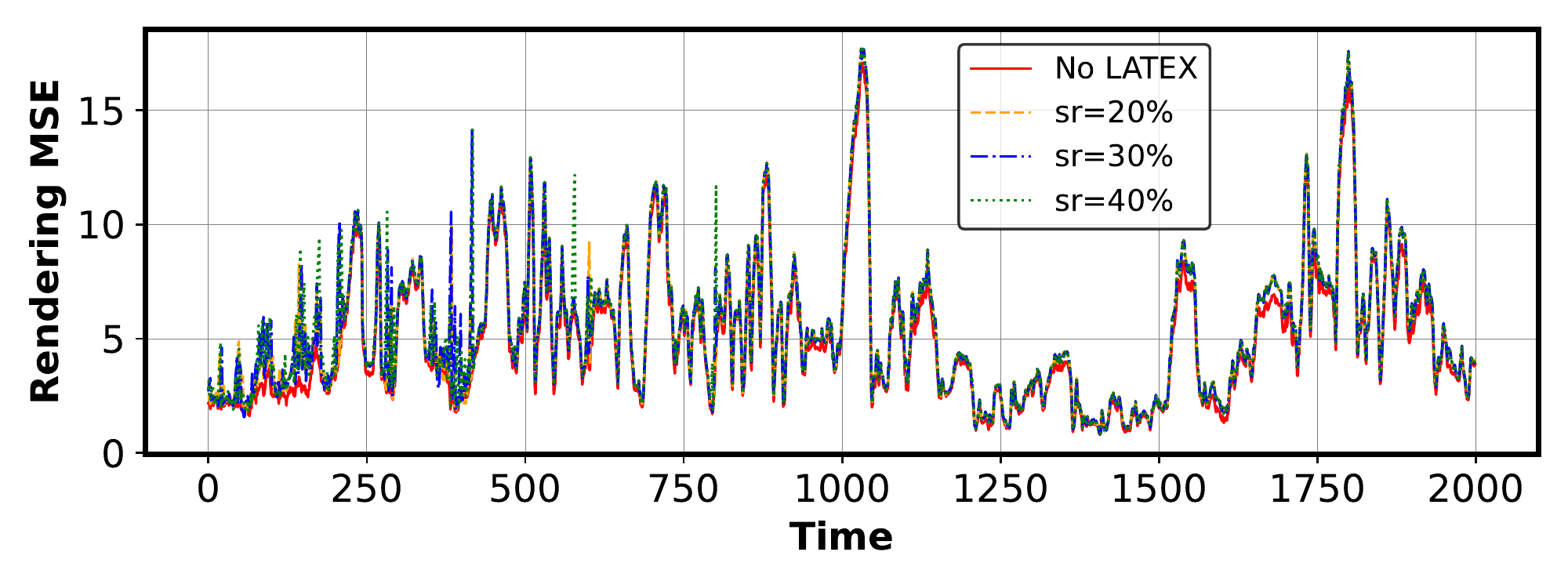}
\vspace{-2.5em}
\caption{Visualize the rendering MSE evolution w/ and w/o LATEX under different skip ratios (denoted as $sr$).}
\label{fig:mse_time}
\vspace{-2em}
\end{figure}

\textbf{Rendering quality.}
To validate the rendering quality of LATEX, we plot the MSE-time of the rendered avatars on the test video of one sampled subject w/o and w/ LATEX under different thresholds, which result in different skip ratios (i.e., the ratio of the frames encoded by linear extrapolation to the total frames). As shown in Fig.~\ref{fig:mse_time}, we observe that our LATEX can skip the encoder inference for 20\%$\sim$30\% frames with a comparable rendering MSE. The detailed LATEX settings are provided in the supplementary material.

\textbf{The overall speed-up.}
We further show the overall speed-up measured on two devices achieved by combining AVE-NAS and LATEX. In particular, we adopt different skip ratios on top of AVE-S and we report the average MSE in the front view across six identities in Tab.~\ref{tab:sota_render_loss} as well as the average latency for encoding one frame measured on two devices. We benchmark with the SOTA EEM encoder and a tiny encoder searched by AVE-NAS in Tab.~\ref{tab:overall_speedup} and we observe that: \underline{(1)} our AVE-NAS can achieve 6.11$\times$/15.39$\times$ speed-up on Quest 2 and Pixel 3, respectively, with comparable rendering quality (+0.08 MSE), and \underline{(2)} enabling both AVE-NAS and LATEX can achieve better latency-MSE trade-offs as compared to enabling AVE-NAS only, e.g., a 1.43$\times$ speed-up on average measured on Quest 2 with a 0.48 MSE reduction over AVE-T. This indicates the necessity of LATEX for ensuring the scalability towards ultra-low-latency/resource settings on AR/VR headsets, which is promising for the distributed near-sensor encoding~\cite{sumbul2022system} in next-generation AR/VR headsets.

\begin{table}[ht]
\vspace{-0.5em}
\centering
\caption{The overall speed-up of combining AVE-NAS and LATEX over EEM and AVE-NAS only, where $sr$ denotes skip ratios.}
\vspace{-0.6em}
\resizebox{0.98\linewidth}{!}
{    
\begin{tabular}{cccccc}
\toprule
\textbf{Model} & \textbf{LATEX} & \textbf{MFLOPs} & \textbf{Quest 2 (ms)} & \textbf{Pixel 3 (ms)} & \textbf{Avg. MSE} \\ \midrule
EEM & - & 2930.77 & 12.48 & 483.47 & 13.68 \\ \midrule
AVE-T & - & 102.60 & 2.37 & 32.40 & 15.72 \\ \midrule
\multirow{4}{*}{AVE-S} & $sr$=0\% & \multirow{4}{*}{174.75} & 2.47 & 37.48 & 13.22 \\
 & $sr$=20\% &  & 2.06 & 31.41 & 13.76 \\
 & $sr$=30\% &  & 1.86 & 28.37 & 14.42 \\
 & $sr$=40\% &  & 1.66 & 25.33 & 15.24 \\ \bottomrule
\end{tabular}
}
\label{tab:overall_speedup}
\vspace{-1.3em}
\end{table}

    \vspace{-0.2em}
\section{Conclusion}
\label{sec:conclusion}

To enable real-time and robust photorealistic avatars on resource-constrained AR/VR headsets, 
in this work, we proposed \METHOD{}, which for the first time achieves real-time and robust driving of Codec Avatars in AR/VR when exclusively using merely on-device computing resources. 
In particular, \METHOD{} minimizes both the model and temporal redundancies via the proposed AVE-NAS and LATEX techniques, respectively, while at the same time enhancing the robustness under extreme expressions. Our method can achieve a 5.05$\times$ speed-up on Meta Quest 2 while maintaining comparable animation quality, and our delivered insights can shed light on future innovations in deploying AR/VR-centric computer vision and neural rendering tasks on AR/VR platforms.

\vspace{-0.3em}
\section*{Acknowledgement}
\vspace{-0.3em}

The work performed by Yonggan Fu and Yingyan (Celine) Lin is supported by a Meta research award and CoCoSys, one of the seven centers in JUMP 2.0, a Semiconductor Research Corporation (SRC) program sponsored by DARPA.

%%%%%%%%% REFERENCES
{\small
\bibliographystyle{ieee_fullname}
\bibliography{egbib}
}

\end{document}

\title{Supplementary Materials of \\ Auto-CARD: Efficient and Robust Codec Avatar Driving \\ for Real-time Mobile Telepresence \vspace{-0.5em}}

\author{Yonggan Fu\textsuperscript{1}, \, Yuecheng Li\textsuperscript{2}, \, Chenghui Li\textsuperscript{2}, \, Jason Saragih\textsuperscript{2}, \\ Peizhao Zhang\textsuperscript{2}, \, Xiaoliang Dai\textsuperscript{2}, \, Yingyan (Celine) Lin\textsuperscript{1} \\
\textsuperscript{1}Georgia Institute of Technology \,
\textsuperscript{2}Meta\\
\small{\{yfu314, celine.lin\}@gatech.edu \, \{yuecheng.li, leo.li, jsaragih, stzpz, xiaoliangdai\}@meta.com} \vspace{-0.5em}
}

\maketitle

\begin{table}[t!]
% \vspace{-0.5em}
\centering
\caption{Benchmark the searched encoders with SOTA encoder designs in terms of measured latency on Quest 2/Pixel 3 and rendering MSE across different 10 identities and 3 view directions.}
\vspace{-0.5em}
\resizebox{0.98\linewidth}{!}
{    
\begin{tabular}{c|cccc|ccc}
\toprule
\multirow{4}[12]{*}{\textbf{Iden.}} & \textbf{Model} & \textbf{EEM} & \textbf{\begin{tabular}[c]{@{}c@{}}EEM\\ -ch50\end{tabular}} & \textbf{\begin{tabular}[c]{@{}c@{}}EEM\\ -res50\end{tabular}} & \textbf{\begin{tabular}[c]{@{}c@{}}AVE-L \\ (Ours)\end{tabular}} & \textbf{\begin{tabular}[c]{@{}c@{}}AVE-M \\ (Ours)\end{tabular}} & \textbf{\begin{tabular}[c]{@{}c@{}}AVE-S \\ (Ours)\end{tabular}} \\ \cmidrule{2-8}
 & \textbf{MFLOPs} & 2930.77 & 765.38 & 747.44 & 605.14 & 306.93 & \textbf{174.75} \\ \cmidrule{2-8}
 & \textbf{\begin{tabular}[c]{@{}c@{}}Lat. (ms)\\ Quest 2\end{tabular}} & 12.48 & 10.02 & 9.40 & 4.59 & 3.26 & \textbf{2.47} \\ \cmidrule{2-8}
 & \textbf{\begin{tabular}[c]{@{}c@{}}Lat. (ms)\\ Pixel 3\end{tabular}} & 483.47 & 164.27 & 117.27 & 70.53 & 52.61 & \textbf{37.78} \\ \midrule
\multirow{3}{*}{\textbf{S1}} & Front & 8.48 & 8.54 & 11.27 & \textbf{6.91} & 7.46 & 7.54 \\
 & Left & 8.11 & 8.36 & 10.84 & \textbf{6.80} & 7.29 & 7.41 \\
 & Right & 8.04 & 8.07 & 10.65 & \textbf{6.45} & 7.01 & 7.03 \\ \midrule
\multirow{3}{*}{\textbf{S2}} & Front & 15.70 & 16.08 & 22.77 & \textbf{14.63} & 15.10 & 16.21 \\
 & Left & 14.52 & 15.08 & 21.08 & \textbf{13.74} & 14.56 & 15.22 \\
 & Right & 17.51 & 18.00 & 24.82 & \textbf{16.17} & 16.31 & 17.53 \\ \midrule
\multirow{3}{*}{\textbf{S3}} & Front & 12.03 & 12.85 & 15.53 & \textbf{10.91} & 11.42 & 12.25 \\
 & Left & 12.00 & 12.88 & 14.96 & \textbf{10.93} & 11.48 & 11.99 \\
 & Right & 12.73 & 13.62 & 16.45 & \textbf{11.41} & 11.83 & 13.05 \\ \midrule
\multirow{3}{*}{\textbf{S4}} & Front & 17.42 & 18.71 & 21.40 & \textbf{15.62} & 16.33 & 16.98 \\
 & Left & 19.12 & 20.41 & 23.22 & \textbf{16.80} & 17.57 & 18.01 \\
 & Right & 17.47 & 18.72 & 21.27 & \textbf{15.56} & 16.08 & 16.94 \\ \midrule
\multirow{3}{*}{\textbf{S5}} & Front & 7.01 & 7.81 & 15.95 & \textbf{5.78} & 5.94 & 6.06 \\
 & Left & 7.32 & 8.05 & 16.09 & \textbf{6.22} & 6.33 & 6.45 \\
 & Right & 7.10 & 7.93 & 15.45 & \textbf{5.92} & 6.03 & 6.24 \\ \midrule
\multirow{3}{*}{\textbf{S6}} & Front & 19.52 & 20.77 & 25.05 & \textbf{17.34} & 18.39 & 19.08 \\
 & Left & 26.28 & 26.47 & 33.76 & \textbf{22.27} & 24.33 & 24.34 \\
 & Right & 15.89 & 16.73 & 22.25 & \textbf{14.27} & 15.41 & 15.55 \\ \midrule
\multirow{3}{*}{\textbf{S7}} & Front & 19.52 & 20.77 & 25.05 & \textbf{17.34} & 19.08 & 18.39 \\
 & Left & 10.03 & 10.23 & 17.00 & \textbf{9.00} & 9.39 & 9.43 \\
 & Right & 10.14 & 10.16 & 18.26 & \textbf{9.02} & 9.31 & 9.54 \\ \midrule
\multirow{3}{*}{\textbf{S8}} & Front & 8.44 & 12.32 & 18.32 & \textbf{6.22} & 6.62 & 7.61 \\
 & Left & 8.94 & 12.98 & 18.57 & 7.19 & \textbf{7.03} & 8.12 \\
 & Right & 8.39 & 12.15 & 18.46 & \textbf{6.55} & 6.61 & 7.56 \\ \midrule
\multirow{3}{*}{\textbf{S9}} & Front & 4.12 & 4.72 & 5.97 & \textbf{3.46} & 3.75 & 3.96 \\
 & Left & 4.46 & 4.61 & 5.82 & \textbf{3.71} & 4.03 & 4.00 \\
 & Right & 6.63 & 7.09 & 9.06 & \textbf{5.31} & 5.83 & 6.15 \\ \midrule
\multirow{3}{*}{\textbf{S10}} & Front & 10.52 & 11.97 & 12.75 & \textbf{9.37} & 9.79 & 9.94 \\
 & Left & 10.38 & 11.75 & 12.40 & \textbf{9.25} & 9.75 & 9.89 \\
 & Right & 11.01 & 12.57 & 12.75 & \textbf{9.89} & 10.18 & 10.33 \\ \midrule
\textbf{Avg.} & - & 11.96 & 13.01 & 17.24 & \textbf{10.47} & 11.01 & 11.43 \\ \bottomrule
\end{tabular}
}
\label{tab:mse}
\vspace{-1em}
\end{table}

\section{More Quantitative Rendering Results}
\label{sec:mse}

\textbf{Rendering MSE across all identities.}
In addition to the 6 identities evaluated in Sec. 5.2 of the main text, we further validate our AVE-NAS searched encoders across another 4 identities and summarize all the rendering MSE in Tab.~\ref{tab:mse}.
We can observe that our searched encoders consistently achieve better MSE-efficiency trade-offs across identities as compared to the SOTA EEM encoder~\cite{schwartz2020eyes} based on the measurement on Meta Quest 2, e.g., our searched AVE-L achieves a 2.72$\times$ speed-up over EEM while also reducing the rendering MSE by 1.49 on average and our AVE-S achieves a 5.05$\times$ speed-up with an average MSE reduction of 0.53.

\textbf{Rendering LPIPS and FID:} We also measure the rendering quality in terms of LPIPS~\cite{zhang2018unreasonable} and FID~\cite{heusel2017gans} (the lower the better), which better aligns with human perception. We benchmark our AVE-L model with the baseline EEM model across six identities. As shown in Tab.~\ref{tab:lpips_fid}, our AVE-L model still outperforms the baseline EEM model, e.g., a 0.39 FID reduction in average with 2.72$\times$ speed-up on Quest 2, indicating that our method can consistently achieve better rendering quality in terms of human perception.

\begin{table}[!h]
\vspace{-0.5em}
\centering
\caption{Benchmark our AVE-L model with the baseline EEM model across six identities in terms of the achieved LPIPS and FID.}
\vspace{-0.5em}
\resizebox{0.98\linewidth}{!}
{    
\begin{tabular}{c|cccc|cccc}
\toprule
\multirow{3}{*}{\textbf{Identity}} & \multicolumn{4}{c|}{\textbf{Front View}} & \multicolumn{4}{c}{\textbf{Left View}} \\ \cmidrule{2-9}
 & \multicolumn{2}{c}{\textbf{LPIPS $\downarrow$}} & \multicolumn{2}{c|}{\textbf{FID $\downarrow$}} & \multicolumn{2}{c}{\textbf{LPIPS $\downarrow$}} & \multicolumn{2}{c}{\textbf{FID $\downarrow$}} \\ \cmidrule{2-9}
 & \textbf{Gabe} & \textbf{Auto-CARD} & \textbf{Gabe} & \textbf{Auto-CARD} & \textbf{Gabe} & \textbf{Auto-CARD} & \textbf{Gabe} & \textbf{Auto-CARD} \\ \midrule
S1 & 0.0740 & \textbf{0.0457} & 4.121 & \textbf{2.506} & 0.0503 & \textbf{0.0459} & 3.382 & \textbf{2.821} \\
S2 & 0.0707 & \textbf{0.0694} & 2.969 & \textbf{2.954} & 0.0713 & \textbf{0.0701} & 3.010 & \textbf{2.864} \\
S3 & 0.0711 & \textbf{0.0667} & 2.260 & \textbf{2.107} & 0.0693 & \textbf{0.0650} & 1.857 & \textbf{1.805} \\
S4 & 0.0669 & \textbf{0.0651} & 2.772 & \textbf{2.730} & 0.0666 & \textbf{0.0630} & 2.570 & \textbf{2.367} \\
S5 & 0.0709 & \textbf{0.0604} & 5.061 & \textbf{4.355} & 0.0754 & \textbf{0.0651} & 3.524 & \textbf{3.119} \\ 
S6 & 0.0707 & \textbf{0.0615} & 3.437 & \textbf{2.930} & 0.0666 & \textbf{0.0620} & 2.869 & \textbf{2.595} \\ \bottomrule
\end{tabular}
}
\label{tab:lpips_fid}
\vspace{-1em}
\end{table}

\begin{table*}[t!]
\vspace{-0.5em}
\centering
\caption{Visualize the operators, channel scales, and input resolutions of searched encoders (zoom-in for better view).}
\vspace{-0.5em}
\resizebox{0.98\linewidth}{!}
{    
\begin{tabular}{c|c|cc|c|c|c|c|c}
\toprule
\multirow{2}{*}{\textbf{Model}} & \multirow{2}{*}{\textbf{\begin{tabular}[c]{@{}c@{}}Input \\ Resolution\end{tabular}}} & \multirow{2}{*}{\textbf{View}} & \multirow{2}{*}{\textbf{\begin{tabular}[c]{@{}c@{}}Searchable\\ Factor\end{tabular}}} & \multirow{2}{*}{\textbf{\begin{tabular}[c]{@{}c@{}}Feature Extraction\\ Backbone\end{tabular}}} & \multicolumn{3}{c|}{\textbf{Branch}} & \multirow{2}{*}{\textbf{\begin{tabular}[c]{@{}c@{}}MFLOPs\\ (Total)\end{tabular}}} \\ \cmidrule{6-8}
 &  &  &  &  & \textbf{Latent Code} & \textbf{Gaze} & \textbf{Key Point} &  \\ \midrule
\multirow{9}{*}{AVE-L} & \multirow{9}{*}{80} & \multirow{3}{*}{Left Eye} & Operator & {[}'fuse-mb', 'conv'{]} & {[}'fuse-mb', 'skip', 'conv', 'conv', 'fuse-mb', 'fuse-mb'{]} & {[}'conv', 'fuse-mb', 'conv', 'fuse-mb', 'conv', 'conv'{]} & {[}'fuse-mb', 'skip'{]} & \multirow{9}{*}{605.14} \\ 
 &  &  & Channel Scale & {[}0.53125, 0.5{]} & {[}0.53125, 0.5, 0.53125, 0.5, 0.75, 0.75{]} & {[}1.0, 0.5, 0.5, 0.5, 0.9375, 0.9375{]} & {[}0.5, 0.5{]} &  \\
 &  &  & MFLOPs & 46.08 & 43.67 & 35.12 & 54.85 &  \\ \cmidrule{3-8}
 &  & \multirow{3}{*}{Right Eye} & Operator & {[}'fuse-mb', 'conv'{]} & {[}'conv', 'skip', 'conv', 'skip', 'conv', 'fuse-mb'{]} & {[}'conv', 'conv', 'conv', 'conv', 'conv', 'fuse-mb'{]} & {[}'fuse-mb', 'skip'{]} &  \\
 &  &  & Channel Scale & {[}0.8125, 0.875{]} & {[}0.5, 0.5, 0.5, 0.5, 0.75, 1.0{]} & {[}1.0, 0.5, 0.6875, 0.5, 0.75, 0.9375{]} & {[}0.5, 0.5{]} &  \\
 &  &  & MFLOPs & 101.12 & 69.92 & 34.13 & 54.85 &  \\ \cmidrule{3-8}
 &  & \multirow{3}{*}{Mouth} & Operator & {[}'fuse-mb', 'fuse-mb'{]} & {[}'conv', 'conv', 'conv', 'fuse-mb', 'fuse-mb', 'fuse-mb'{]} & - & - &  \\
 &  &  & Channel Scale & {[}0.53125, 0.5{]} & {[}0.5, 0.5, 0.53125, 0.53125, 1.0, 1.0{]} & - & - &  \\
 &  &  & MFLOPs & 34.77 & 129.43 & - & - &  \\ \midrule \midrule
\multirow{9}{*}{AVE-M} & \multirow{9}{*}{64} & \multirow{3}{*}{Left Eye} & Operator & {[}'fuse-mb', 'conv'{]} & {[}'fuse-mb', 'skip', 'skip', 'skip', 'conv', 'conv'{]} & {[}'fuse-mb', 'conv', 'conv', 'fuse-mb', 'fuse-mb', 'conv'{]} & {[}'skip', 'skip'{]} & \multirow{9}{*}{306.93} \\
 &  &  & Channel Scale & {[}0.53125, 0.5{]} & {[}0.5, 0.5, 0.5, 0.5, 0.75, 0.75{]} & {[}1.0, 0.5, 0.5, 0.5, 1.0, 1.0{]} & {[}0.5, 0.5{]} &  \\
 &  &  & MFLOPs & 38.88 & 31.35 & 24.84 & 5.31 &  \\   \cmidrule{3-8}
 &  & \multirow{3}{*}{Right Eye} & Operator & {[}'conv', 'conv'{]} & {[}'skip', 'skip', 'conv', 'skip', 'conv', 'conv'{]} & {[}'conv', 'conv', 'conv', 'fuse-mb', 'fuse-mb', 'fuse-mb'{]} & {[}'skip', 'skip'{]} &  \\
 &  &  & Channel Scale & {[}0.5, 0.5{]} & {[}0.5, 0.5, 0.5, 0.5, 1.0, 1.0{]} & {[}0.5, 0.5, 0.5, 0.5, 0.875, 0.9375{]} & {[}0.5, 0.5{]} &  \\
 &  &  & MFLOPs & 37.67 & 57.96 & 16.09 & 5.31 &  \\ \cmidrule{3-8}
 &  & \multirow{3}{*}{Mouth} & Operator & {[}'conv', 'skip'{]} & {[}'conv', 'skip', 'conv', 'conv', 'conv', 'fuse-mb'{]} & - & - &  \\
 &  &  & Channel Scale & {[}0.5, 0.5{]} & {[}0.5, 0.5, 0.53125, 0.5, 1.0, 1.0{]} & - & - &  \\
 &  &  & MFLOPs & 28.07 & 60.25 & - & - &  \\ \midrule \midrule
\multirow{9}{*}{AVE-S} & \multirow{9}{*}{64} & \multirow{3}{*}{Left Eye} & Operator & {[}'conv', 'skip'{]} & {[}'skip', 'skip', 'skip', 'skip', 'fuse-mb', 'conv'{]} & {[}'fuse-mb', 'skip', 'conv', 'fuse-mb', 'conv', 'conv'{]} & {[}'skip', 'skip'{]} & \multirow{9}{*}{174.75} \\
 &  &  & Channel Scale & {[}0.5, 0.53125{]} & {[}0.5, 0.5, 0.5, 0.5, 0.75, 0.75{]} & {[}0.5625, 0.5, 0.53125, 0.5, 0.75, 0.9375{]} & {[}0.5, 0.53125{]} &  \\
 &  &  & MFLOPs & 28.07 & 10.91 & 6.14 & 5.31 &  \\ \cmidrule{3-8}
 &  & \multirow{3}{*}{Right Eye} & Operator & {[}'conv', 'skip'{]} & {[}'skip', 'skip', 'skip', 'skip', 'conv', 'fuse-mb'{]} & {[}'fuse-mb', 'skip', 'conv', 'conv', 'conv', 'fuse-mb'{]} & {[}'skip', 'skip'{]} &  \\
 &  &  & Channel Scale & {[}0.53125, 0.53125{]} & {[}0.5, 0.5, 0.5, 0.5, 0.75, 1.0{]} & {[}0.5625, 0.5, 0.5, 0.5, 0.75, 0.9375{]} & {[}0.5, 0.53125{]} &  \\
 &  &  & MFLOPs & 28.07 & 14.65 & 6.82 & 5.31 &  \\ \cmidrule{3-8}
 &  & \multirow{3}{*}{Mouth} & Operator & {[}'conv', 'skip'{]} & {[}'fuse-mb', 'skip', 'conv', 'conv', 'conv', 'fuse-mb'{]} & - & - &  \\
 &  &  & Channel Scale & {[}0.5, 0.5{]} & {[}0.5, 0.5, 0.5, 0.5, 0.75, 1.0{]} & - & - &  \\
 &  &  & MFLOPs & 28.07 & 40.21 & - & - &  \\ \bottomrule
\end{tabular}
}
\label{tab:searched_encoders}
\vspace{-1em}
\end{table*}

\section{Detailed Experiment Setup}
\label{sec:setup}

\textbf{Training settings.} We follow the same setting in~\cite{Gabriel20} to train the decoders, which is fixed during encoder search and training. For encoder search and training, we adopt the same training schedules for architecture parameters and model weights. In particular, we search/train the encoder for 50K/100K steps, respectively, using an Adam optimizer with a batch size of 16 and an initial learning rate of 1e-3 decayed by 0.1 every 40K steps. 
The weighted coefficients for $\mathcal{L}_{latent}$, $\mathcal{L}_{gaze}$, $\mathcal{L}_{geo}$,$\mathcal{L}_{tex}$, $\mathcal{L}_{kpt}$, and $\mathcal{L}_{ren}$ in Eq. (4) of the main text are set to 1e-1, 1, 1, 1, 1e3, and 1e-4, respectively, for balancing their magnitudes.
The training of the whole Codec Avatar pipeline\cite{Gabriel20} takes about 12 GPU days and the search and training of avatar encoders take 2 GPU days.

\textbf{Hyper-parameters of our proposed \METHOD{}.} 
\underline{Setup of AVE-NAS:}
For the sampling frequency $K$ for resolution search in Sec. 4.2.3, we set it as 16, which can help stabilize the resolution search process based on our empirical experiments. For the search objective of AVE-NAS, we set $\tau$ and $m$ in Eq. (3) of the main text, which are the temperature parameter and the momentum factor, respectively, as 10 and 0.9 across all the experiments. 
\underline{Setup of LATEX:}
For LATEX introduced in Sec. 4.3.2 of the main text, we extrapolate the latent codes based on the previous 4 frames ($T$=4). To avoid error accumulation across frames caused by inaccurate extrapolation, we also set a rule that if the latent codes of the previous 3 frames are derived by LATEX, we enforce an encoder inference for acquiring the latent code of the current frame. In addition, LATEX's lightweight prediction head, featuring one convolutional layer, one pooling layer, and one fully-connected layer, is applied after the convolutional head in the feature extraction backbone (introduced in Sec.~\ref{sec:architecture}) to early predict the latent code, which is used to adaptively decide whether to perform extrapolation via comparing with a threshold determined by the target skip ratio.

\section{Visualize the Encoder Architectures}
\label{sec:architecture}

\textbf{SOTA EEM encoder.} The SOTA EEM encoder~\cite{Gabriel20}, which requires 2.9 GFLOPs to encoder each frame, adopts a feature extractor backbone to extract the features of HMC captured images, on top of which a latent code branch, a gaze branch, and a key point branch,  are applied to estimate the latent code, gaze, and key points of the current frame, respectively. In particular, the input resolution is [192, 192, 3] which concatenates the captured infrared images of the left eye/right eye/mouth views. The feature extractor backbone is composed of  one convolution head and two bottleneck blocks in~\cite{he2016deep} with an output channel number of [64, 128, 128], respectively, and a total stride of 4; The latent code branch comprises 6 bottleneck blocks featuring a total stride of 8 and an output channel number of [128, 128, 256, 256, 256, 256], respectively, and one fully-connected layer to map the latent features into a 256-d vector; The gaze branch comprises 6 bottleneck blocks featuring a total stride of 8 and an output channel number of [128, 128, 128, 128, 64, 64], respectively, plus one pooling layer and a fully-connected layer to map the gaze features to a 6-d vector (i.e., 3-d gaze directions for each eye); The key point branch comprises 4 bottleneck blocks with an output channel number of [128, 128, 64, 64], respectively, followed by a convolutional layer to regress 19 key points.

\textbf{Search space of our AVE-NAS.} To ensure sufficient flexibility of the encoder architecture, the search space spans operator types, depth, width, and input resolution. For each view in the view-decoupled supernet, we set 2, 6, 6, 2 searchable blocks for the feature extractor backbone, latent code branch, gaze branch, and key point branch with a maximal output channel of [64,64], [64, 64, 256, 256, 512,512], [256, 256, 128, 128, 32, 32], and [128, 64], respectively. For the supported operator types, each searchable block can be set as  Fused-MBConv~\cite{tan2021efficientnetv2}, a single convolution with a kernel size of 3$\times$3, or skip connections. Note that if there exists a feature dimension mismatch, the skip connection is switched to a convolutional layer with a kernel size of 1$\times$1. The encoder depth is searchable via setting the operators to skip connections and the searchable encoder width is achieved by applying an output channel scale, which is searched from [0.5, 0.53125, 0.5625, 0.59375, 0.625, 0.6875, 0.75, 0.8125, 0.875, 0.9375, 1.0], on top of the maximal channel numbers. For input resolution search, the candidates are [32, 48, 64, 80, 96, 128, 192] where the maximally allowed resolution is inherited from EEM~\cite{Gabriel20}.

\textbf{Our searched encoders.} We visualize the searched encoders, including AVE-L, AVE-M, and AVE-S, in Tab.~\ref{tab:searched_encoders}. We can observe that the searched branches for different views feature diverse complexity, aligning with the fact that different cameras can capture different aspects of facial appearance and motion. This indicates that the view-decoupled supernet design not only echos the future distributed encoding trend, but also enables the flexibility of view-specific encoder structures and thus improves the achievable accuracy-efficiency trade-off of avatar encoding.

{\small
\bibliographystyle{ieee_fullname}
\bibliography{egbib}
}